\documentclass[arxiv]{melba}

\usepackage{mwe} 

\usepackage{amsmath,amsfonts}

\usepackage{soul}
\usepackage{xurl}

\melbaid{YYYY:NNN}  
\doi{10.59275/j.melba.2024-AAAA}
\melbaauthors{Benveniste}  
\email{pierre-louis-2.benveniste@polymtl.ca}
\volume{3}
\firstpageno{1337}  
\melbayear{2026}  
\datesubmitted{yyyy-m1-d1}  
\datepublished{yyyy-m2-d2}  

\ShortHeadings{Multiple sclerosis lesion tracking}{Benveniste}

\title{Longitudinal tracking of multiple sclerosis lesions in the spinal cord: A validation study}

\author{
	\firstname Pierre-Louis \surname Benveniste\aff{1},
    \firstname Julian \surname McGinnis\aff{2},
	\firstname Shannon \surname Kolind\aff{3},
    \firstname Larry D. \surname Lynd\aff{3},
    \firstname Sarah A. \surname Morrow\aff{4},
    \firstname Jiwon \surname Oh\aff{5},
    \firstname Alexandre \surname Prat\aff{6},
    \firstname Alice \surname Schabas\aff{3},
    \firstname Penelope \surname Smyth\aff{7},
    \firstname Roger \surname Tam\aff{3},
    \firstname Anthony \surname Traboulsee\aff{3},
    \firstname Mark \surname Mühlau\aff{2}
    \firstname Herve \surname Lombaert\aff{1},
    \firstname Julien \surname Cohen-Adad\aff{1},
}

\affiliations{
	\num 1 \addr Polytechnique Montreal, Montreal, QC, Canada \\
    \num 2 \addr TUM University Hospital, Munich, Germany \\
	\num 3 \addr University of British Columbia, Vancouver, BC, Canada \\
	\num 4 \addr University of Calgary, Calgary, AB, Canada \\
    \num 5 \addr University of Toronto, Toronto, ON, Canada \\
    \num 6 \addr Université de Montréal, Montreal, QC, Canada \\
    \num 7 \addr University of Alberta, Edmonton, AB, Canada \\
}

\abstract{
	Longitudinal characterization of multiple sclerosis (MS) lesions remains constrained by the lack of frameworks capable of establishing consistent instance-level correspondences across time. Conventional segmentation approaches produce semantic lesion masks at each visit and therefore fail to capture the complex instance temporal patterns associated with lesion appearance, disappearance, splitting, or merging. This study presents a comparative evaluation of five strategies for automated tracking of spinal cord MS lesions in longitudinal MRI data from a multi-site cohort. The investigated strategies rely either on deformable registration or on a spinal anatomical reference system, and encompass overlap-based matching, coordinate-based Hungarian algorithm, gradient-boosted classification, and Siamese model classification. Tracking accuracy is quantified using instance-level true positives, false positives, and false negatives, allowing to assess the presence of one-to-many and many-to-one associations. Results show best performance for the registration-based overlap method. This study provides the first systematic analysis of lesion-instance correspondence in the spinal cord and outlines the strengths and limitations of registration-based and registration-free paradigms for longitudinal MS assessment. The code is available at: \url{https://github.com/ivadomed/longitudinal-sc-ms-lesion-tracking}}


\begin{document}

\twocolumn[\maketitle]

\section{Introduction}
	\enluminure{M}{ultiple sclerosis} (MS) is an inflammatory disease of the central nervous system that produces demyelinating lesions in the brain and spinal cord (SC). Longitudinal monitoring of these lesions with magnetic resonance imaging (MRI) is essential to detect change in disease course and adapt treatment appropriately, for example, in cases of conversion from relapsing-remitting to secondary progressive \citep{Krieger2016-ur}. In routine practice, however, assessment of subtle lesion change remains approximate, i.e., there are limited measures of lesion growth and radiomic features. Instance lesion segmentation offers the opportunity to provide quantitative measures (e.g., lesion morphometrics, radiomic extraction), thereby improving clinical interpretation \citep{Lipp2020-pf}. 

    Substantial progress has been made in segmentation of brain \citep{Dereskewicz2025-yh, Kaur2021-uk, Wiltgen2024-cp, Cerri2023-vh} and SC MS lesions \citep{naga2025automatic, Benveniste2025-cq, benveniste2026}. Typically, these segmentations are encoded as a single-class value and are used to produce biomarkers such as total lesion load. However, instance-level metrics are needed to characterize specific lesions of interest, such as “critical” lesions \citep{Ahmad2025-th}. In addition, monitoring how each lesion evolves over time adds prognosis value. Achieving this not only requires accurate lesion delineation at each time point, but also precise tracking of lesions across longitudinal scans, that is, establishing a correspondence between lesion instances across multiple time points \citep{Santoro-Fernandes2024-uo}. 
    
    Although several longitudinal lesion segmentation frameworks have been proposed for the brain \citep{Carass2017-ht,Kamraoui2022-vi,Rokuss2025-qa, Kruger2020-yx}, they do not provide a consistent lesion-instance correspondence and thus cannot be used for tracking purposes. To date, there is no dedicated methodology for the tracking of MS lesions in the SC. Only one study has explored lesion tracking in the brain, utilizing registered scans and a non-zero overlap criterion to track lesions \citep{Kohler2019-yo}. In contrast, object tracking has been extensively explored in other domains, typically relying on one-to-one assignments derived from spatial overlap, trajectory regularity, or bipartite matching techniques such as the Hungarian algorithm \citep{Bolme2010-oa,Li2019-mr,Li2018-ej,Rokuss2025-nw,Hering2021-jp}. Only a limited subset of medical imaging studies \citep{Santoro-Fernandes2024-uo,Di_Veroli2023-xt} has considered asymmetric matching, addressing the more challenging one-to-many or many-to-one correspondences arising from lesion appearing, resolving, splitting and merging events \citep{Lassmann2018-mb}.

\begin{figure}[h]
		\centering
		\includegraphics[width=0.9\linewidth]{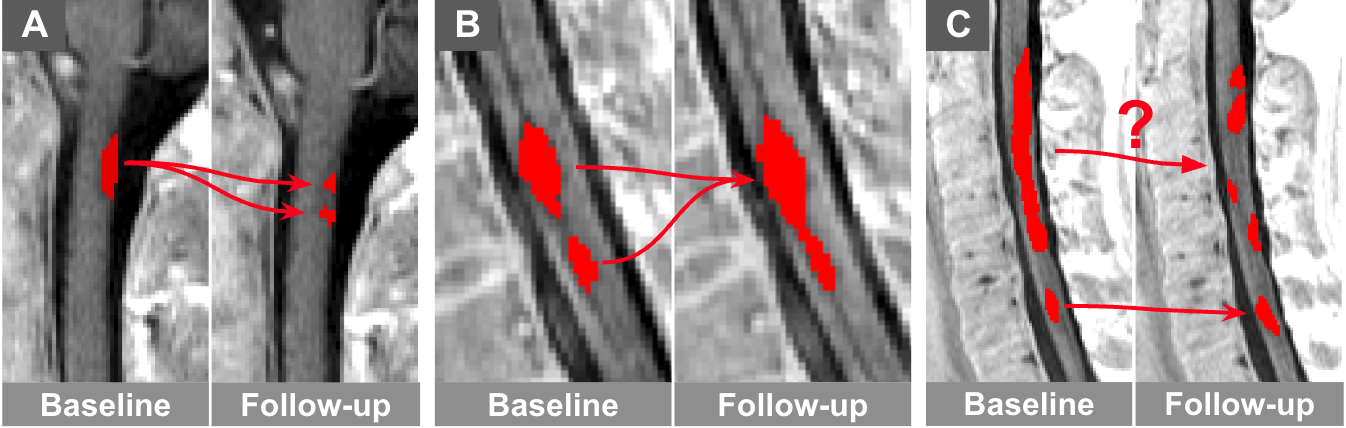}
		\caption{Examples of one–to–many, many–to–one, and uncertain longitudinal correspondences in SC MS lesions. A: A single focal lesion at baseline divides into two distinct lesions at follow-up, illustrating a one-to-many evolution. B: Two focal lesions at baseline merge, creating a confluent lesion at follow-up. C: A diffuse lesion is segmented as a single confluent region at baseline but is delineated as several smaller lesions at follow-up. That last example highlights how variability in the segmentation can produce markedly different longitudinal representations of the same underlying pathology.}
        \label{fig:complex-corres}
	\end{figure}

    To contextualize these complex lesion temporal evolutions, Figure \ref{fig:complex-corres}A illustrates a case interpreted by the rater as a splitting event, while Figure \ref{fig:complex-corres}B shows a merging event. Diffusely abnormal white matter, extensively studied in both brain \citep{Musall2024-cm} and SC imaging \citep{Bergers2002-fn}, may be segmented as a single confluent region or as multiple smaller clusters, depending on rater's interpretation. Figure \ref{fig:complex-corres}C shows an example where a diffuse lesion was delineated differently across visits. These scenarios expose fundamental limitations of existing tracking techniques \citep{Hering2021-jp,Tang2022-im, Rokuss2025-nw,Yan_undated-pv}, which cannot deal with asymmetric correspondences.

    Tracking SC lesions longitudinally introduces additional specific methodological challenges. While longitudinal brain alignment traditionally relies on rigid or affine registration, SC alignment often requires non-linear registration due to the cord’s inherently articulated structure \citep{Commowick2012-hh, Beal2023-zz}. Moreover, the SC is a thin structure ($\sim$1cm diameter), and MRI-visible lesions are small and susceptible to partial-volume effects. Even minimal misregistration can substantially alter lesion boundaries. This raises concerns regarding the reliability of registration-based approaches for temporal correspondence. Lesion tracking is further complicated by the pronounced inter- and intra-rater variability in SC lesion delineation, with reported fluctuations in lesion count reaching up to 50\% \citep{Walsh2023-to}. Such variability reinforces the inadequacy of one-to-one correspondence assumptions and highlights the need for methods capable of handling uncertain or evolving lesion morphology.

    In this work, we evaluate "standard" tracking methods leveraging registration and the Hungarian algorithm. We also investigate more complex methods using anatomical-based coordinate systems and deep learning methods to handle the complex, asymmetric temporal correspondence of evolving lesions. The objective of this work is to validate multiple automated frameworks for the complex tracking of SC MS lesion instances across longitudinal MRI using multi-site (n=5) clinical datasets.

\section{Related Works}
	Research on longitudinal lesion analysis in MS has predominantly focused on the brain, notably pushed in 2017 by the MS longitudinal lesion segmentation challenge \citep{Carass2017-ht}. In this challenge, all scans were registered to a baseline scan, and human raters showed poor longitudinal consistency when blinded to time-order of scan. Top-performing methods relied on convolutional neural networks or classical machine-learning models. Following this initiative, several deep learning frameworks were proposed. Kamraoui et al. \citep{Kamraoui2022-vi} introduced a Siamese encoder coupled with a decoder trained to segment lesions from two affine-registered time points. Rokuss et al \citep{Rokuss2024-qe} used dual encoders and feature-difference weighting blocks to emphasize temporal changes. An additional encoder–decoder formulation operating on pre-registered scans was described in \citep{Kruger2020-yx}. All these approaches rely critically on image registration, which, according to Diaz-Hurtado et al. \citep{Diaz-Hurtado2022-ms}, remains a mandatory component in all brain-based longitudinal pipelines. Importantly, these methods focus exclusively on longitudinal segmentation and do not perform lesion tracking. To date, no work has addressed longitudinal segmentation or tracking of MS lesions in the SC.

    Beyond segmentation, longitudinal instance tracking has been investigated in several medical and non-medical settings, although not in MS. In computer vision, object tracking has been extensively studied, with high-performing methods leveraging Siamese networks, correlation filters, and others \citep{Bolme2010-oa,Li2018-ej,Li2019-mr}. These models are typically optimized for 2D, single-object trajectories, stable appearance characteristics, and minimal topological change, conditions that are incompatible with the evolution of MS lesions. In medical imaging, lesion tracking has mainly focused on tumor evolution. LesionLocator \citep{Rokuss2025-nw} propagates a bounding box through time to track a tumor, implicitly assuming that the target remains a single connected entity. The DeepLesion dataset \citep{Yan_undated-pv} has supported the development of various approaches, including transformer-based architectures for temporal propagation \citep{Tang2022-im} and registration-based matching strategies \citep{Hering2021-jp}. These frameworks generally assume one-to-one correspondences across time points and are therefore unable to capture more complex patterns such as lesion splitting or merging.
    
    Only a few studies have explicitly addressed asymmetric correspondences. Santoro-Fernandes et al. proposed a cancer lesion tracking method \citep{Santoro-Fernandes2024-uo}. To deal with lesion merging, they perform lesion clustering on dilated lesion masks using geometrical considerations such as lesion distance and longest lesion chord. Lesion groups are matched pairwise across scans using the Hungarian algorithm. Di Veroli et al. also investigated cancer lesion tracking \citep{Di_Veroli2023-xt} by matching lesions using deformable registration and overlap. Lesion segmentations were dilated to compensate for small registration errors. Their overlap threshold was set at 10\% and tracking performance was evaluated using true positives, false positives, and false negatives.
    
    Overall, existing longitudinal MS segmentation frameworks do not perform lesion tracking. Only a handful of tracking methods in other domains can handle one-to-many and many-to-one correspondences. The specific challenges posed by SC MS lesions (small size, high morphological variability, rater variability, and dependence on non-rigid registration) underscore the need to develop tracking frameworks that can explicitly model complex lesion evolution, potentially leveraging deep learning to improve temporal consistency.

\section{Methods}
	\subsection{Dataset}

    The study used longitudinal data from the CanProCo cohort \citep{Oh2021-ol}, each scanned at baseline (M0) and 12-month follow-up (M12). All participants with at least one lesion in the SC, for whom manual longitudinal lesion tracking had been completed, were included in this study (n=34, 13 males and 21 females, age range at M0: 18-59 y.o.). Data were acquired across 5 hospitals (Calgary, Edmonton, Montreal, Toronto, and Vancouver) on Philips, Siemens, or GE systems, using sagittal PSIR (n=29) or STIR (n=5) sequence, at 0.7×0.7×3 mm resolution. While upstream lesion and disc segmentation methods used isotropic resampled images at inference \citep{benveniste2026,Warszawer2024-wa}, the final masks are mapped back into the native raw space. Consequently, tracking and registration steps are performed on native-space masks.

    The datasets used in this research work are not publicly available; however, for reproducibility purposes, we provide the binary segmentations, vertebral disc levels, and registration warping fields in the GitHub release (version r20260610).
    
    \subsection{Lesion segmentation}
    
    Lesion segmentations were generated using \textit{sct\_deepseg lesion\_ms} \citep{benveniste2026} available in SCT \citep{De_Leener2017-ka} v7.2. The model is a 5-fold U-Net optimized using the nnUNet framework. Inference was performed using a 5-fold ensemble with test-time augmentation based on mirroring along all axes. 
    
    Instance-level masks were extracted by connected component analysis with 26-voxel connectivity.
    
    \subsection{Data split}

    The dataset was split on a participant basis into training (n=25) and testing (n=9) sets. For each participant, all lesion pairs across M0 and M12 were enumerated. Each pair was labeled as positive if it represented the same longitudinal lesion instance, and negative otherwise. The resulting dataset was composed of 579 pairs for training (n=91 positive pairs) and 134 pairs for testing (n=24 positive pairs). 
    
    \subsection{Lesion tracking}

    To thoroughly investigate the longitudinal tracking of SC lesions, we strategically selected five methods that span two key axes of variation: registration-based versus registration-free tracking, and one-to-one versus asymmetric matching scheme. Strategies \#1 and \#4 deploy the classical Hungarian algorithm to establish rigid, one-to-one assignments. Strategies \#2 and \#3 are registration-free, learning-based approaches that operate in the SC anatomical coordinate system. Strategy \#5 relies on voxel-level overlap after deformable registration, which naturally accommodates asymmetric one-to-many and many-to-one correspondences, but is expected to be sensitive to registration errors and segmentation noise.
    
    \subsubsection{Strategy \#1: Hungarian algorithm in the SC coordinate system}
    
    This approach defined each lesion solely by its centroid in (\textit{z}, \textit{r}, $\theta$) coordinates, defined in the SC coordinate system.
    
\begin{figure}[h]
		\centering
		\includegraphics[width=0.7\linewidth]{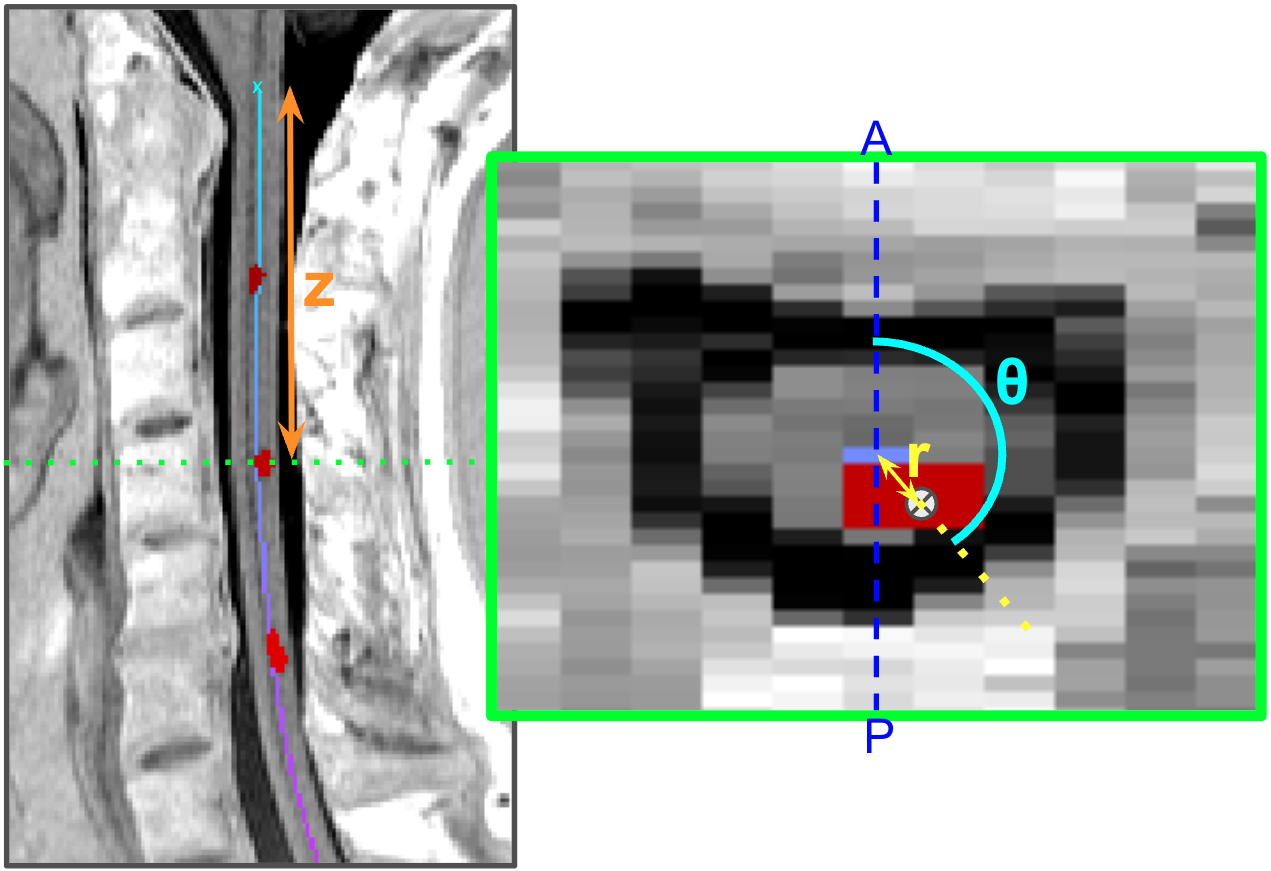}
		\caption{Illustration of cylindrical coordinates defining lesion centroids in the SC anatomical system. Left: sagittal PSIR image with lesions (in red) and the labeled centerline (in blue to pink). Right: axial slice corresponding to the green dotted line, where $\otimes$  denotes the centroid of the lesion. A: Anterior, P: Posterior}
        \label{fig:sc-coord}
	\end{figure}

    In the SC coordinate system (as detailed in Figure \ref{fig:sc-coord}), lesion centroids were expressed in cylindrical anatomical coordinates : \textit{z}: continuous vertebral position along the labeled centerline obtained from \textit{sct\_get\_centerline} and disc levels from TotalSpineSeg \citep{Warszawer2024-wa}; \textit{r}: radial distance from the centerline; $\theta$: angular orientation relative to the anterior–posterior axis. These coordinates enabled registration-free comparison of lesion locations across time.

    Longitudinal correspondences between M0 and M12 lesions were obtained using the Hungarian algorithm \citep{Kuhn1955-vq} applied to Euclidean distances between centroids. In brief, the Hungarian algorithm computes the optimal one-to-one assignment between two sets by minimizing the total assignment cost. To reflect the higher spatial reliability of the \textit{z} axis in sagittal acquisitions, distances were anisotropically weighted with a 20:1 ratio between the \textit{z} axis and the transverse plane. This ratio was selected on the train set from the following set [5, 10, 15, ..., 50]. This ratio ensures that vertebral-level alignment is prioritized over transverse proximity, preventing incorrect matches between lesions at different cord levels that might otherwise appear close in cylindrical coordinates.
    
    \subsubsection{Strategy \#2: XGBoost using lesion features in the SC coordinate system}
    
    This strategy formulated lesion tracking as a supervised binary classification task. An XGBoost model \citep{Chen2016-si} was trained to discriminate matching and non-matching pairs of lesions from both time points of the same participants, using geometric descriptors. Each lesion was described by its SC-coordinate centroid (\textit{z}, \textit{r}, and $\theta$), volume, and maximum Right-Left (RL)/Anterior-Posterior (AP)/Inferior-Superior (IS) diameters. Pair-level descriptors included component-wise displacements, Euclidean distance, and volume difference.
    
    Model training used a log-loss objective. Hyperparameter optimization was performed using Bayesian optimization over 50 iterations with three-fold cross-validation and the average-precision score as the selection criterion. The final configuration retained after optimization was: colsample\_bytree = 0.7998, learning\_rate = 0.0555, max\_depth = 7, min\_child\_weight = 7, scale\_pos\_weight = 9, and subsample = 0.8585.
    
    Lesion pairs were assigned to the same tracks when their predicted probability exceeded 0.5. This strategy circumvented the one-to-one constraint inherent to Hungarian matching and avoided explicit registration steps.
    
    \subsubsection{Strategy \#3: Siamese network using lesion features in the SC coordinate system}
    
    A Siamese model \citep{Dey2017-dr} was trained to discriminate lesion pairs of the same track. Each lesion was described by its SC-coordinate centroid (\textit{z}, \textit{r}, and $\theta$), volume, and maximum RL/AP/IS diameters. Features were standardized prior to training using z-score normalization.
    
    Both encoder branches consisted of a dense encoder (128 → 64 → 32) with ReLU activation and batch normalization. The resulting embeddings were compared using an L2 distance, followed by a sigmoid unit yielding a match probability. Training used a contrastive loss with margin 1, optimized via RMSprop (learning rate of 0.001). Because matching pairs were rare, batch-level class weights were used (0.59 for negatives, 3.28 for positives). Training proceeded for 500 epochs with a batch size of 20.
    
    \subsubsection{Strategy \#4: Hungarian matching in the baseline scan reference space}
    
    Follow-up segmentations were first warped to the baseline image (M0) using \textit{sct\_register\_multimodal} (parameters: $step=0,type=label,dof=Tx\_Ty\_Tz; step=1,type=im,algo=dl$)  from SCT \citep{De_Leener2017-ka}. The registration method is based on a first step of pre-alignment of scans using disc levels and a second step performing registration using a deep learning model \citep{Beal2023-zz}. Intervertebral disc labeling, required for initialization, was performed using TotalSpineSeg \citep{Warszawer2024-wa}. All registrations underwent visual quality control to characterize the robustness of the method and identify the root causes of any potential failure cases. No registrations were manually adjusted or excluded based on this quality control, thereby maintaining the fully automated nature of the results presented in this study. Lesion centroids were computed in warped segmentation masks.
    
    Lesion centroids were then computed in the baseline space, and correspondences were determined through Hungarian matching based on centroid distances.
    
    \subsubsection{Strategy \#5: Lesion overlapping in the baseline scan reference space}
    
    Follow-up segmentations were warped to the baseline scan space using the same registration method as in Strategy \#4. This strategy paired lesions across time points when their intersection-over-union (IoU) was non-zero. This voxel-overlap rule provided a direct, registration-based criterion for lesion continuity.
    
    \subsection{Evaluation metrics}
    
\begin{figure}[h]
		\centering
		\includegraphics[width=0.9\linewidth]{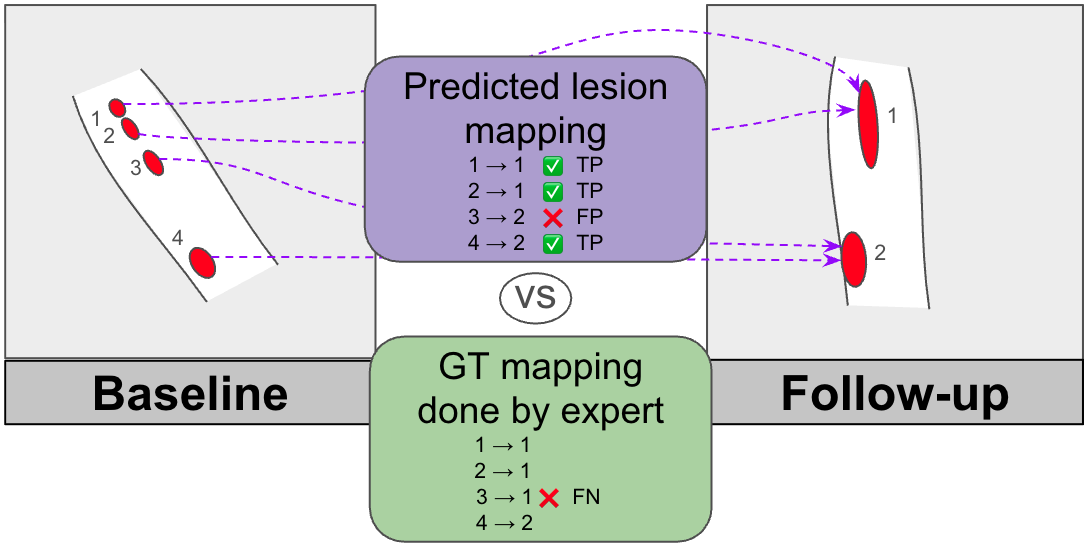}
		\caption{Evaluation of predicted lesion mappings. Predicted lesion mappings (purple box) are compared to GT lesion mappings done by an experts (green box). In this example, three lesions were correcly matched (TP=3), one lesion mapping was incorrect (FP=1) and one lesion mapping was missing (FN=1).}
        \label{fig:eval-fig}
	\end{figure}

    Tracking performance was quantified by comparing the predicted lesion mappings with the ground-truth (GT) longitudinal mappings (Figure \ref{fig:eval-fig}). To establish the mappings, expert raters had access to both the raw MRI scans at baseline and follow-up, as well as the corresponding lesion segmentations. To specifically address the challenges of identifying lesion splitting, merging, or morphological evolution, raters also had access to the co-registered images and warped segmentations.

    To accurately reflect the complexity of MS lesion evolution, we utilized the following link-based definitions:
    \begin{itemize}
        \item True Positives (TP): the number of correctly identified follow-up lesions belonging to the same GT longitudinal track. Because a single baseline lesion may evolve into multiple follow-up lesions, TP may exceed 1 for a given baseline lesion. For example, if a baseline lesion splits into two and the tracker correctly identifies both, it is counted as two TPs.
        \item False Positives (FP): the number of follow-up lesions incorrectly associated with a baseline lesion.
        \item False Negatives (FN): the number of missing follow-up assignments. These occur when a baseline lesion is not associated with its corresponding follow-up lesion(s). In cases of lesion splitting, if only one of two follow-up lesions is correctly mapped, the missing link is recorded as a FN. 
    \end{itemize}

    A baseline lesion that has fully resolved by follow-up, or a lesion appearing at follow-up with no baseline counterpart, each contributes one FP per incorrectly predicted link — that is, only if a tracking strategy erroneously associates them with a lesion from the other timepoint. If left unmatched, they contribute neither FP nor FN.
    
    These metrics follow the same principles adopted in previous work on MS lesion tracking with one-to-many and many-to-one trajectories \citep{Di_Veroli2023-xt,Santoro-Fernandes2024-uo} and provide an interpretable basis for quantifying correct, spurious, and missed longitudinal correspondences.

    To address the limited size of the test set (n=9) and provide more reliable performance estimates across the full cohort, we conducted a subject-wise Leave-One-Out Cross-Validation (LOOCV, n=34 iterations) on all 34 participants for all five strategies. In each iteration, one participant was held out for evaluation while the remaining 33 were used for training and hyperparameter selection. For Strategy \#1, the optimal anisotropic weighting ratio was selected from the candidate set [5, 10, 15, ... 50] by maximizing the F1-score on the 33 training participants, and then applied to the held-out participant. For Strategies \#2 and \#3, their respective training procedures were performed on the 33 training participants and the resulting models evaluated on the held-out participant. Strategies \#4 and \#5 involve no trainable parameters or hyperparameter selection; their deterministic registration-and-matching pipelines were applied directly to each held-out participant.

    It should be noted that link-level precision and recall, when aggregated across all subjects, give proportionally higher weight to participants with a larger number of lesion pairs, which could bias results toward over-linking. To mitigate this, the LOOCV analysis (Table~\ref{tab:perf_loocv}) computes metrics independently per participant before averaging, ensuring that each participant contributes equally to the final performance estimate regardless of lesion burden. An additional experiment computing metrics per participant was also computed for Strategy \#5 (Appendix \ref{app:per_participant_results} Table \ref{tab:average_participant_metrics}).

\section{Results}

\begin{figure*}[h]
		\centering
		\includegraphics[width=0.8\linewidth]{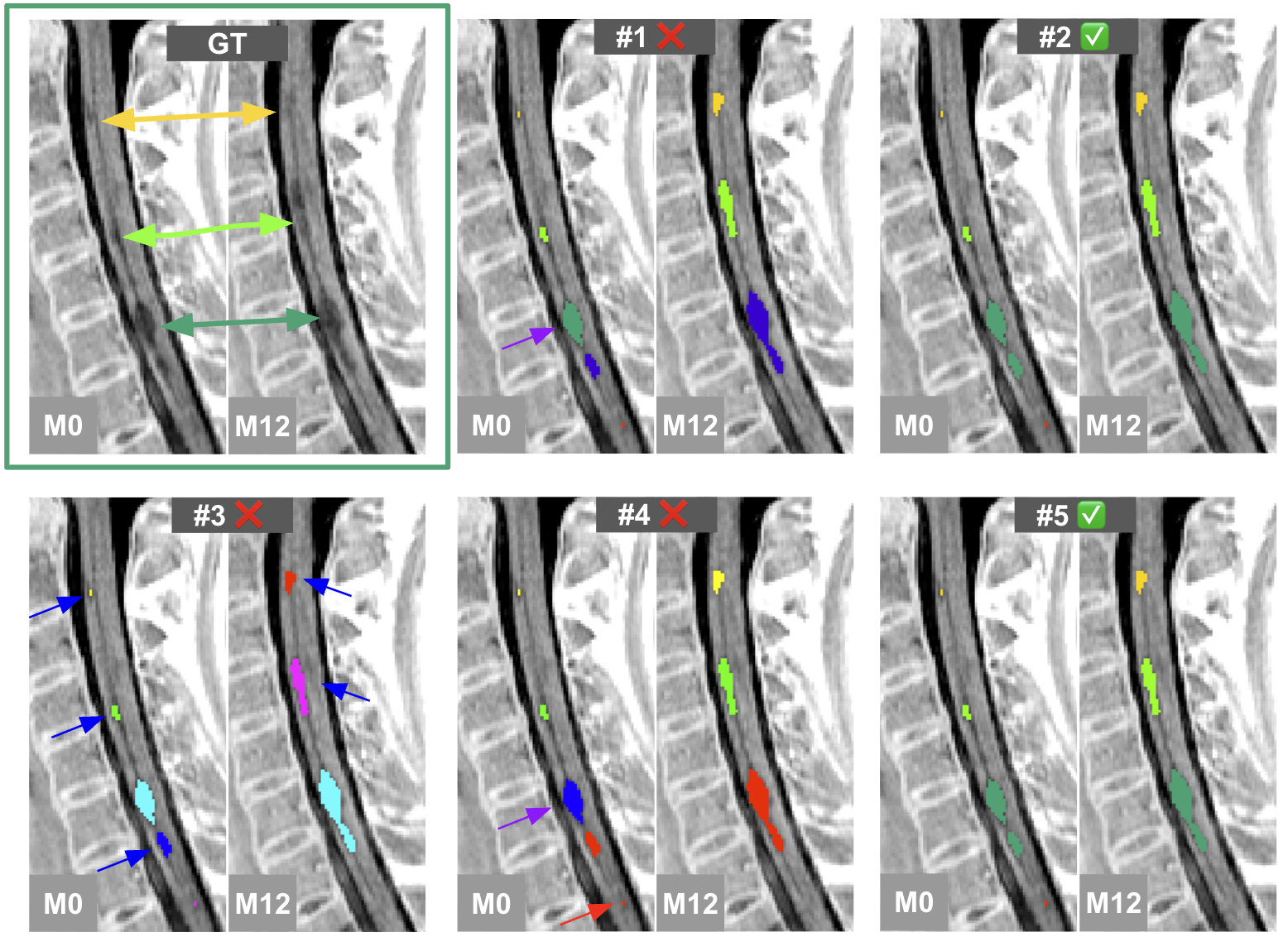}
		\caption{Qualitative comparison of the five tracking strategies on a PSIR scan from the Toronto site. Each image shows labeled lesion instances at M0 and M12, where lesions sharing the same color represent a predicted longitudinal trajectory. GT represents the correct lesion association between the two time points. \textcolor{red}{$\rightarrow$}: false-positive mapping; \textcolor{violet}{$\rightarrow$}: incorrect correspondence; \textcolor{blue}{$\rightarrow$}: false-negative mapping.}
        \label{fig:quali-result}
	\end{figure*}

    Figure \ref{fig:quali-result} compares the five tracking strategies in a challenging case with a high lesion burden. In this example, Strategies \#2 and \#5 yielded accurate longitudinal correspondences, whereas Strategy \#3 exhibited substantial mismatches. Strategies \#1 and \#4 were inherently constrained by the one-to-one assignment imposed by the Hungarian algorithm, limiting their ability to recover the many-to-one trajectory in this example.

    Longitudinal tracking performance for each strategy was quantified using predicted lesion segmentations (Table \ref{tab:test_results}). Overall, Strategy \#5, combining registration to baseline with IoU-based correspondence, achieved the highest performance, reaching an F1 score of 0.98 on the test set. Results on all the data splits are detailed in Table \ref{tab:all_results} in Appendix \ref{app:all_results}. Per-site and per-participant results are available in Appendix \ref{app:per_site_results} and Appendix \ref{app:per_participant_results}.
    
    \begin{table}[h] 
		\centering
		\caption{Longitudinal lesion-tracking performance across the five strategies on the test set \underline{on predicted lesion segmentations}.}
        \begin{tabular}{l|c|c|c|c|c}
               & \#1 & \#2 & \#3 & \#4 & \#5 \\
              \hline
              TP $\uparrow$ & 21 & 21 & 11 & 21 & 23\\
              FP $\downarrow$ & 5 & 1 & 1& 5 & 0\\
              FN $\downarrow$ & 3 & 3 & 13 & 3 & 1\\
              Precision$\uparrow$ & 0.81 & 0.95 & 0.92& 0.81 & 1.00\\
              Recall $\uparrow$ & 0.88 & 0.88 & 0.46 & 0.88 & 0.96\\
              F1 score $\uparrow$ & 0.84 & 0.91 & 0.61& 0.84 & 0.98\\
        \end{tabular}
        \label{tab:test_results}
	\end{table}
    
    To compare the top-performing strategies, we conducted a statistical analysis using the Wilcoxon signed-rank test. On the independent test set (n=9), the performance difference between the registration-based overlap (Strategy \#5) and the registration-free XGBoost model (Strategy \#2) was not statistically significant ($p > 0.05$ for all metrics). 

    To evaluate the stability of these tracking strategies independently of the segmentation model bias, we repeated the entire evaluation framework using manual expert segmentations as the initial lesion masks (Table \ref{tab:perf_manual_seg_test}). When evaluated on these manual masks, Strategy \#5 maintained its superior performance, achieving a test-set F1-score of 0.91, while Strategy \#2 remained highly competitive with an F1-score of 0.81. Although all strategies exhibited a slight reduction in scores, the overall performance hierarchy remained identical. Performances on the other data split are available in Appendix \ref{app:perf_manual_seg} Table \ref{tab:perf_manual_seg_all}.

    \begin{table}[h] 
		\centering
		\caption{Longitudinal lesion-tracking performance on the test set across the five strategies evaluated \underline{on manual lesion segmentations}.}
        \begin{tabular}{l|c|c|c|c|c}
               & \#1 & \#2 & \#3 & \#4 & \#5 \\
              \hline
              TP $\uparrow$ & 21 & 30& 9& 16& 29\\
              FP $\downarrow$ & 8& 9& 4& 9& 0\\
              FN $\downarrow$ & 14& 5& 26& 19& 6\\
              Precision$\uparrow$ & 0.72& 0.77& 0.69& 0.64& 1.00\\
              Recall $\uparrow$ & 0.60& 0.86& 0.26& 0.46& 0.83\\
              F1 score $\uparrow$ & 0.66& 0.81& 0.38& 0.53& 0.91\\
        \end{tabular}
        \label{tab:perf_manual_seg_test}
	\end{table}
    
    To provide robust, symmetric performance estimates across all five strategies, we conducted a subject-wise LOOCV (n=34 iterations) across all five strategies simultaneously. Pairwise Wilcoxon signed-rank tests with Holm-Bonferroni correction were used to compare strategies. Mean LOOCV performance is summarized in Table \ref{tab:perf_loocv}. Strategy \#5 achieved the highest performance across all metrics, and significantly outperformed all other strategies in Precision and F1 ($p<0.05$), with the exception of the F1 comparison against Strategy \#2 ($p=0.062$).

    \begin{table}[h] 
		\centering
		\caption{Longitudinal lesion-tracking performance using LOOCV across the five strategies evaluated \underline{on predicted lesion segmentations}. (* $p<0.05$ for Wilcoxon signed-rank tests with Holm-Bonferroni correction)}
        \begin{tabular}{l|c|c|c|c|c}
               & \#1 & \#2 & \#3 & \#4 & \textbf{\#5} \\
              \hline
              Precision$\uparrow$ & 0.87& 0.92& 0.49& 0.88& \textbf{1.00}*\\
              Recall $\uparrow$ & 0.89& 0.93& 0.35& 0.89& 0.95\\
              F1 score $\uparrow$ & 0.87& 0.91& 0.39& 0.87& 0.97\\
        \end{tabular}
        \label{tab:perf_loocv}
	\end{table}

    It should be noted that the current cohort size is borderline for non-parametric significance testing; future studies with larger samples may provide the statistical power necessary to further disentangle the strengths of these two paradigms.
    
\begin{figure*}[h]
		\centering
		\includegraphics[width=0.9\linewidth]{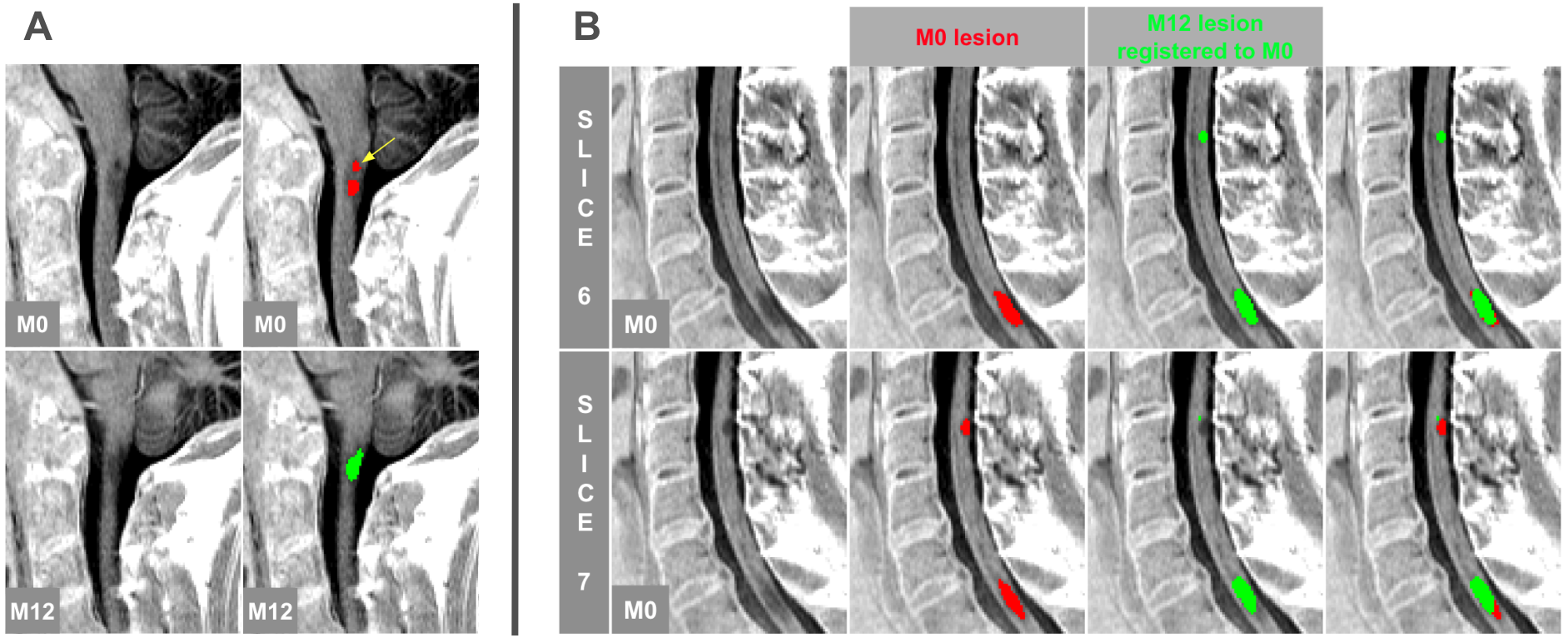}
		\caption{Representative failure cases of Strategy \#5 for longitudinal lesion tracking. (A) False-negative correspondence caused by under-segmentation at baseline, where a lesion is split into two components; the superior component (yellow arrow) fails to overlap with the follow-up lesion, partly due to a slight rightward inter-timepoint shift. (B) Missed correspondence induced by subtle registration errors, with the baseline lesion appearing on slice 6 and the follow-up lesion on slice 7, compounded by a small inferior–superior misalignment.}
        \label{fig:fail_seg}
	\end{figure*}

    The limitations of Strategy \#5 primarily stem from imperfections in the automatically generated segmentations and inaccuracies in inter-timepoint registration. Figure \ref{fig:fail_seg} illustrates two representative failure cases:
    \begin{itemize}
        \item \textbf{Segmentation noise:} In Figure \ref{fig:fail_seg}A, a baseline lesion that should have been represented as a single contiguous structure was instead segmented into two smaller components, leading to a false negative during tracking. In particular, the superior component (indicated by the yellow arrow) was not associated with the follow-up lesion because no spatial overlap was observed. This lack of overlap was exacerbated by a slight rightward shift between the sagittal acquisitions at the two time points.
        \item \textbf{Small Registration/Partial Volume Shifts:} In Figure \ref{fig:fail_seg}B, lesion correspondence failed due to subtle right-left misregistration: the baseline lesion was visible on slice 6, whereas the corresponding follow-up lesion appeared on slice 7, preventing successful matching under the overlap-based criterion. This discrepancy was further amplified by a small residual misalignment along the inferior–superior axis.
    \end{itemize}

    To facilitate reuse, the XGBoost and Siamese models from Strategies \#2 and \#3, trained on the full dataset (train and test), are made available in the GitHub release (version r20260610).
    
\section{Discussion}
    
    In this study, we assessed five strategies for the longitudinal tracking of MS lesions in the SC. The comparison of tracking approaches underscores the intrinsic difficulty of defining consistent lesion correspondences in the SC, where anatomical deformability, small and irregularly shaped lesions, substantial inter-rater variability, and frequent lesion merging or division complicate temporal association.
    
    Among the evaluated strategies, the registration-based IoU method (\#5) achieved the highest tracking performance. Allowing non-zero overlap proved advantageous for handling large volumetric changes, diffuse confluent lesions, or small registration errors. While higher thresholds are common in general object tracking, they are unsuitable for MS lesions, which can undergo drastic volumetric changes or merge into diffuse confluent regions or divide into smaller lesions. As illustrated in Figure \ref{fig:complex-corres}C, a higher threshold would likely result in false negatives when matching a baseline large diffuse lesion to a significantly smaller follow-up instance. A sensitivity analysis of Strategy \#5 to the IoU threshold (Appendix \ref{app:iou_thresh_exp} Table \ref{tab:iou_sweep}) confirms that the non-zero overlap criterion is optimal for this application and robust for any threshold below 0.01. Precision remains perfect across all tested thresholds, while Recall degrades monotonically from 0.96 to 0.42 as the threshold increases from near-zero to 0.5. This confirms that higher thresholds, common in general object tracking, are ill-suited for MS lesions, which frequently undergo drastic volumetric changes, splitting, or merging between timepoints. Nonetheless, IoU remains an imperfect criterion: it is sensitive to segmentation noise, deformations introduced by registration, and substantial longitudinal lesion evolution. As illustrated in Figure \ref{fig:quali-result}, subtle differences in segmentation can propagate into substantial discrepancies in downstream correspondence assignments. Crucially, the superior performance of Strategy \#5 should not be interpreted only as a virtue of the tracking algorithm, but also as a direct proxy for the high fidelity of the underlying inter-timepoint registration pipeline. While feature-based approaches like Strategy \#2 rely on learned geometric features to actively compute tracking choices, Strategy \#5 succeeds primarily because the co-registration aligns the anatomy well enough to render complex matching logic redundant.

    There is no established practice for subject-specific longitudinal analysis in SC MRI. Previous studies registered follow-up images to the baseline, without using a template-based approach \citep{Salem2018-yo, Salem2020-yq}. Regarding the potential bias introduced by the choice of M0 as the fixed reference frame, we note that for Strategy \#5, this choice is mathematically inconsequential: the IoU criterion is symmetric by definition ($IoU(A,B) = IoU(B,A)$). Strategy \#4, which computes lesion centroids in the warped follow-up image, should be the most sensitive to this choice, as interpolation and resampling errors introduced during registration are not perfectly symmetric. As shown in Appendix \ref{app:reg_direction} Table \ref{tab:reg_direction}, the performance of both strategies is minimally affected by registration direction, i.e., by performing registration from the baseline to the follow-up (M0 → 12) or from the follow-up to the baseline (M12 → M0). F1 scores remain stable at 0.83-0.85 for Strategy \#4 and 0.96 for Strategy \#5, regardless of direction. The small observed differences for Strategy \#5 are consistent with voxel-level interpolation artifacts introduced by the warping field, which can marginally alter lesion boundaries, causing artificial splitting and/or merging.
    
    As illustrated in Figure \ref{fig:fail_seg}A, minor errors in lesion segmentation can significantly impact subsequent tracking performance. It is important to note that the segmentation model employed here was rigorously validated in prior work, which demonstrated that automated predictions were comparable and in some cases superior to manual annotations \citep{benveniste2026}. Nevertheless, our sensitivity analysis using manual segmentations in Table \ref{tab:perf_manual_seg_test} confirms that Strategies \#5 (IoU-based) and \#2 (XGBoost) consistently achieve the highest performance, regardless of whether the initial segmentations were generated manually or automatically.
    
    The performance achieved in this study, particularly the F1-score of 0.98 for Strategy \#5, is highly competitive when compared to state-of-the-art tracking in other medical domains. Santoro-Fernandes et al. reported an F1-score of 0.88 for automated whole-body lesion tracking in oncology from 1570 lesion pairs from 8 patients \citep{Santoro-Fernandes2024-uo}. Furthermore, our results are consistent with the high Precision (0.91–0.98) and Recall (0.87–0.99) reported by Di Veroli et al. for longitudinal CT lesion tracking on 83 scans from 19 patients \citep{Di_Veroli2023-xt}. The slightly higher performance observed in our study may be attributed to the relatively small duration between baseline and follow-up scans (12 months). These comparisons suggest that while SC MS lesion tracking presents unique technical difficulties, the use of specialized registration and anatomical coordinate systems allows for highly accurate longitudinal monitoring.
    
    Strategies relying on bipartite assignment via the Hungarian algorithm (\#1 and \#4) were structurally constrained, as they enforce one-to-one mappings. This assumption is incompatible with the one-to-many and many-to-one trajectories commonly observed in MS. Their lower performance reflects this inherent limitation. Despite this constraint, the coordinate-based Hungarian approach (\#1) performed surprisingly well. This suggests that the proposed anatomical coordinate system captures meaningful spatial structure that can support tracking.
    
    The LOOCV analysis corroborates and strengthens the conclusions drawn from the held-out test set. Strategy \#5 retained the highest overall performance (F1=0.97), achieving perfect precision across all 34 folds, and significantly outperforming all other strategies in Precision and F1 ($p<0.05$), with the sole exception of the F1 comparison against Strategy \#2 ($p=0.062$). Strategy \#2 remained the strongest registration-free alternative (F1=0.91), with no statistically significant difference from Strategy \#5 in either Recall ($p=0.758$) or F1 ($p=0.062$), confirming that geometric descriptors alone can encode salient longitudinal information without deformable registration. Its tendency toward over-confident predictions, reflected in higher false positives than Strategy \#5, may indicate that the model captures patterns associated with early-stage lesion merging. The stability of Strategy \#2 confirmed through LOOCV highlights its potential as a reliable registration-free alternative, especially in cases where non-rigid registration may be computationally expensive or prone to failure. Interestingly, the anisotropic weighting ratio for Strategy \#1 consistently converged to a value of 20 across all 34 LOOCV folds, suggesting that this ratio is a stable and generalizable property of the SC coordinate system rather than an artifact of a particular data split.

    The Siamese architecture (\#3), despite extensive experimentation—including alternative normalization schemes, deeper or wider encoders, different distance metrics, multiple loss formulations, and a range of optimizers—did not yield gains beyond simpler models. It remains unclear whether this underperformance stems primarily from the limited size of the training set or from the restricted discriminative power of the input features. A deliberate methodological choice in this study was to restrict tracking criteria to low-dimensional spatial and geometric descriptors in the SC coordinate system, avoiding the use of lesion shape or appearance from the MRI scans. By focusing on geometric descriptors, we aimed to ensure the method remains independent of MRI contrast and acquisition site, thereby promoting generalizability. Subsequent validation on larger, multi-contrast datasets will be necessary to evaluate whether the Siamese model's performance was primarily constrained by the data volume or if the current feature set is inherently insufficient for this type of architecture. 
    
    This study relied on a disc-based SC reference system, in which vertebral levels provide an indirect surrogate for local SC anatomy. This approach remains sensitive to head and neck positioning differences across sessions: changes in cervical curvature can shift the cord relative to the vertebrae, thereby biasing coordinate estimates. Recent developments using the pontomedullary junction as a reference \citep{Bedard2023-ty} offer a promising alternative by anchoring coordinates directly to intrinsic SC anatomy.
    
    The reliability of all five strategies is intrinsically tied to the accuracy of the labeling of intervertebral discs. This step is required both for the definition of the SC coordinate system (Strategies \#1-3) and for the initialization of the registration pipeline (Strategies \#4–5). While a failure in disc labeling would render the subsequent tracking invalid, no such failures were observed in the current cohort. This performance aligns with prior literature on the TotalSpineSeg algorithm \citep{Warszawer2024-wa}, which demonstrates a robust disc labeling accuracy of 0.99. Nevertheless, in clinical scenarios involving extreme pathology or severely limited fields of view, the robustness of the entire tracking framework would remain contingent on the success of this initial anatomical localization.

    While native-resolution MRI scans were utilized in this study, a potential avenue to mitigate tracking failures induced by subtle spatial misalignments is the adoption of isotropic resampling or super-resolution frameworks prior to temporal pairing. As observed in the failure case illustrated in Figure 5B, the highly anisotropic native resolution ($0.7\times0.7\times3\text{ mm}$) leaves the tracking pipeline vulnerable to sub-voxel shifts in the slice direction, which can entirely eliminate voxel overlap for smaller lesions. Although our current tracking pipeline retains the native raw space to avoid interpolation artifacts, executing the co-registration on isotropically resampled or super-resolved volumes could restore spatial continuity across adjacent slices and prevent tracking failures.
    
    A persistent challenge highlighted in this work is the inadequacy of existing instance-level evaluation metrics for assessing longitudinal lesion tracking. Metrics commonly used in instance segmentation, such as Difference in Count, Mean Precision, Mean Coverage Loss, or Average Best Overlap \citep{Molina2025-fl}, do not capture temporal correspondence. Similarly, multi-object tracking metrics (MOTA, MOTP) \citep{Bernardin2008-qr} are inappropriate because they either quantify segmentation quality or assume strictly one-to-one temporal mappings. Even the Panoptic Quality metric used in recent MS lesion instance segmentation work \citep{Wynen2024-lb} does not extend naturally to longitudinal association. This underscores the need for dedicated metrics tailored to tracking in cases of complex correspondences.
    
    The proposed lesion tracking strategy relies on establishing pairwise correspondence between lesions across adjacent time points, making it naturally scalable to studies with more than two longitudinal scans. Extending to a greater number of follow-ups involves composing these pairwise correspondences sequentially, without altering the fundamental tracking criteria. The evaluation framework, being link-based, rather than focused on the entire trajectory, would remain valid in multi-timepoint scenarios by independently assessing the correctness of each temporal association. Nevertheless, while the pairwise approach is highly scalable, more sophisticated methods could be developed to fully leverage the entire longitudinal sequence simultaneously. Future research should explore alternative, "sequence-aware" methods or architectures. By analyzing all available time points concurrently, these methods could potentially achieve superior performance by capturing non-linear evolution patterns or resolving ambiguities that a simple chain of pairwise links might overlook.
    
    Alternative definitions of lesion instances more complex than connected components, such as Confluent Lesion Splitting (CLS), have been proposed to better represent confluent lesions \citep{Wynen2024-lb,Lou2021-km}. While such approaches might conceptually align with the challenges of longitudinal tracking, prior evidence \citep{Wynen2024-lb} indicates that CLS can degrade segmentation quality relative to connected components, suggesting a trade-off between anatomical realism and segmentation accuracy.

    Beyond the tracking of SC MS lesions, this study offers a generalizable framework for longitudinal instance tracking across broader medical imaging domains. The optimal tracking paradigm depends on two primary variables: registration quality and topological mapping complexity. Where high-quality registration is achievable, a simple overlap rule (Strategy \#5) natively and effectively accommodates asymmetric one-to-many or many-to-one mappings. Conversely, when registration fidelity is compromised, registration-free, feature-based models (Strategy \#2) should be favored; by utilizing low-dimensional, contrast-agnostic geometric descriptors, these frameworks capture splitting or merging events without propagating alignment errors. Ultimately, this validation demonstrates that moving away from rigid one-to-one assignments toward flexible, asymmetric matching scheme is essential whenever target pathologies undergo morphological fission or fusion over time.

\section{Conclusion}

    This study provided the first investigation of longitudinal instance-level tracking of MS lesions in the SC. By formulating lesion tracking as a correspondence problem rather than a segmentation task, the work addressed a gap unfilled by existing longitudinal MS frameworks. In particular, we address the difficult problem of one-to-many and many-to-one associations in object tracking. The results demonstrated that registration-based overlap yielded the highest overall performance, indicating that spatial alignment remains a powerful—although imperfect—mechanism for temporal lesion association. Yet, the competitive performance of the SC–coordinate–based XGBoost model showed that meaningful longitudinal information can be extracted without deformable registration, suggesting an opportunity for registration-free approaches. Future research should validate these findings in larger cohorts and explore the impact of tracking lesions individually.


\acks{JCA received funding from the Canada Research Chair in Quantitative Magnetic Resonance Imaging [CRC-2020-00179], the Canadian Institute of Health Research [PJT-190258, PJT-203803], the Canada Foundation for Innovation [32454, 34824], the Fonds de Recherche du Québec - Santé [322736, 324636], the Natural Sciences and Engineering Research Council of Canada [RGPIN-2019-07244], the Canada First Research Excellence Fund (IVADO and TransMedTech), the Courtois NeuroMod project, the Quebec BioImaging Network [5886, 35450], INSPIRED (Spinal Research, UK; Wings for Life, Austria; Craig H. Neilsen Foundation, USA), Mila - Tech Transfer Funding Program, the Bavarian State Ministry for Science and Art (Collaborative Bilateral Research Program Bavaria – Quebec: AI in medicine, grant F.4-V0134.K5.1/86/34), and Praxis Institute. MM received funding from the German Research Foundation, DFG Priority Programme 2177, Radiomics-Next Generation of Biomedical Imaging (grant 428223038). SK received funding from MS Canada. PLB received funding from the Fonds de recherche du Québec Nature et technologies [370582], UNIQUE Excellence Scholarship and the Canada Research Chair in Shape Analysis in Medical Imaging. JO received funding from the Waugh Family Chair in MS Research at the University of Toronto, CanProCo funding from MS Canada, Brain Canada, Biogen-Idec, Roche.}

%
\ethics{The work follows appropriate ethical standards in conducting research and writing the manuscript, following all applicable laws and regulations regarding treatment of animals or human subjects.}

\coi{Penelope Smyth has received speaker honoraria and participated on advisory boards for EMD Serono Canada, Novartis Pharmaceuticals, Roche Pharmaceuticals, Amgen Canada Pharmaceuticals. Mark Mühlau has received speaker honoraria from Merck. Jiwon Oh has received research funding from Biogen-Idec and Roche (relevant to content of this work).}

\data{The datasets used in this research work are not publicly
available; however, for reproducibility purposes, we provide the binary segmentations, vertebral disc levels and registration warping fields in the GitHub release (\url{https://github.com/ivadomed/sc-ms-lesion-tracking/releases/tag/r20260610}).}

\bibliography{melba-samplebibliography}


\clearpage

\onecolumn

\appendix

\section{Results on all the data splits}\label{app:all_results}
    
    \begin{table*}[h]
        \centering
        \caption{Longitudinal lesion-tracking performance across the five strategies \underline{on predicted lesion segmentations}.}\label{tab:all_results}
        \begin{tabular}{l|cc|cc|cc|cc|cc}
            \bfseries & \multicolumn{2}{c|}{\#1} & \multicolumn{2}{c|}{\#2} & \multicolumn{2}{c|}{\#3} & \multicolumn{2}{c|}{\#4} & \multicolumn{2}{c}{\#5}\\
            & Train& Test & Train& Test & Train& Test & Train& Test & Train& Test\\
            \hline
            TP $\uparrow$ & 76& 21& -- & 21& -- & 11 & 75& 21& 84& 23\\
            FP $\downarrow$ & 21& 5& -- & 1& -- & 1& 16& 5& 0& 0\\
            FN $\downarrow$ & 16& 3& -- & 3& -- & 13& 16& 3& 7& 1\\
            Precision$\uparrow$ & 0.78& 0.81& -- & 0.95& -- & 0.92& 0.82& 0.81& 1.00& 1.00\\
            Recall $\uparrow$ & 0.83& 0.88& -- & 0.88& -- & 0.46& 0.82& 0.88& 0.92& 0.96\\
            F1 score $\uparrow$ & 0.80& 0.84& -- & 0.91& -- & 0.61& 0.82& 0.84& 0.96& 0.98\\
      \end{tabular}  
    \end{table*}

\section{Per-site results of Strategy \#5}\label{app:per_site_results}

    \begin{table*}[h]
    \centering
    \caption{Per-site lesion tracking performance for Strategy \#5.}
    \label{tab:per_site_performance}
    \begin{tabular}{lcccccc}
        \hline
        & Calgary & Edmonton & Montreal & Toronto (test set) & Vancouver \\
        & (n=5) & (n=14) & (n=3) & (n=9) & (n=3) \\
        \hline
        TP$\uparrow$         & 13 & 37 & 16 & 23 & 18 \\
        FP$\downarrow$       & 0  & 0  & 0  & 0  & 0  \\
        FN$\downarrow$       & 1  & 4  & 0  & 1  & 2  \\
        Precision$\uparrow$  & 1.00 & 1.00 & 1.00 & 1.00 & 1.00 \\
        Recall$\uparrow$     & 0.93 & 0.90 & 1.00 & 0.96 & 0.90 \\
        F1-score$\uparrow$   & 0.96 & 0.95 & 1.00 & 0.98 & 0.95 \\
        \hline
    \end{tabular}
\end{table*}

\section{Per-participant results of Strategy \#5}\label{app:per_participant_results}

    \begin{table*}[h]
    \centering
    \caption{Per-participant average lesion tracking performance for Strategy \#5.}
    \label{tab:average_participant_metrics}
    \begin{tabular}{lccc}
        \hline
        & All participants & Train set & Test set \\
        \hline
        Avg. TP$\uparrow$    & 3.15 & 3.36 & 2.56 \\
        Avg. FP$\downarrow$  & 0.00 & 0.00 & 0.00 \\
        Avg. FN$\downarrow$  & 0.24 & 0.28 & 0.11 \\
        Precision$\uparrow$  & 1.00 & 1.00 & 1.00 \\
        Recall$\uparrow$     & 0.95 & 0.94 & 0.98 \\
        F1-score$\uparrow$   & 0.97 & 0.96 & 0.99 \\
        \hline
    \end{tabular}
\end{table*}

\section{Lesion tracking evaluation on manual segmentations}\label{app:perf_manual_seg}

    \begin{table*}[h]
        \centering
        \caption{Longitudinal lesion-tracking performance across the five strategies evaluated \underline{on manual lesion segmentation}.}
        \label{tab:perf_manual_seg_all}
        \begin{tabular}{l|cc|cc|cc|cc|cc}
             & \multicolumn{2}{c|}{\#1} & \multicolumn{2}{c|}{\#2} & \multicolumn{2}{c|}{\#3} & \multicolumn{2}{c|}{\#4} & \multicolumn{2}{c}{\#5} \\
            Metric & Train& Test & Train& Test & Train& Test & Train& Test & Train& Test \\
            \hline
            TP $\uparrow$        & 85& 21 & - & 30 & - & 9& 45& 16 & 97& 29 \\
            FP $\downarrow$     & 22&  8 &  - &  9 &  - &  4& 28&  9 &  4&  0 \\
            FN $\downarrow$     & 21& 14 &   - &  5 & - & 26& 61& 19 &  9&  6 \\
            Precision $\uparrow$& 0.79& 0.72 & - & 0.77 & - & 0.69& 0.62& 0.64 & 0.96& 1.00 \\
            Recall $\uparrow$   & 0.80& 0.60 & - & 0.86 & - & 0.26& 0.42& 0.46 & 0.92& 0.83 \\
            F1-score $\uparrow$ & 0.79& 0.66 & - & 0.81 & - & 0.38& 0.50& 0.53 & 0.94& 0.91 \\
        \end{tabular}
    \end{table*}

\newpage

\section{Sensitivity of Strategy \#5 to IoU threshold}\label{app:iou_thresh_exp}

\begin{table}[h]
    \centering
    \caption{Sensitivity of Strategy \#5 to IoU threshold on the test set.}
    \begin{tabular}{l|c|c|c|c|c|c|c|c}
        & $\leq$0.01 & 0.05 & 0.10 & 0.15--0.20 & 0.25 & 0.30--0.35 & 0.40 & 0.45--0.50 \\
        \hline
        TP$\uparrow$        & 23 & 22 & 20 & 18 & 17 & 13 & 12 & 10 \\
        FP$\downarrow$      & 0  & 0  & 0  & 0  & 0  & 0  & 0  & 0  \\
        FN$\downarrow$      & 1  & 2  & 4  & 6  & 7  & 11 & 12 & 14 \\
        Precision$\uparrow$ & \textbf{1.00} & 1.00 & 1.00 & 1.00 & 1.00 & 1.00 & 1.00 & 1.00 \\
        Recall$\uparrow$    & \textbf{0.96} & 0.92 & 0.83 & 0.75 & 0.71 & 0.54 & 0.50 & 0.42 \\
        F1$\uparrow$        & \textbf{0.98} & 0.96 & 0.91 & 0.86 & 0.83 & 0.70 & 0.67 & 0.59 \\
    \end{tabular}
    \label{tab:iou_sweep}
\end{table}

\section{Effect of registration direction on registration-based strategies}\label{app:reg_direction}

\begin{table}[h]
    \centering
    \caption{Effect of registration direction on the performance of Strategies \#4 
    and \#5 on the predicted lesion segmentations of the entire dataset.}
    \begin{tabular}{l|c|c||c|c}
        & \multicolumn{2}{c||}{\#4} & \multicolumn{2}{c}{\#5} \\
        & M12$\rightarrow$M0 & M0$\rightarrow$M12 & M12$\rightarrow$M0 & M0$\rightarrow$M12 \\
        \hline
        TP$\uparrow$         & 96 & 99 & 107 & 108 \\
        FP$\downarrow$       & 21 & 19 & 0 & 1 \\
        FN$\downarrow$       & 19 & 16 & 8 & 7 \\
        Precision$\uparrow$  & 0.82 & 0.84 & 1.00 & 0.99 \\
        Recall$\uparrow$     & 0.83 & 0.86 & 0.93 & 0.94\\
        F1$\uparrow$         & 0.83 & 0.85 & 0.96 & 0.96\\
    \end{tabular}
    \label{tab:reg_direction}
\end{table}

\end{document}